\documentclass[10pt,twocolumn,letterpaper]{article}

\usepackage{iccv}              

\DeclareMathOperator*{\argmax}{arg\,max}

\usepackage{amsmath}
\usepackage{amssymb}
\usepackage{booktabs}
\usepackage[accsupp]{axessibility}  

\usepackage{ntheorem}

\newtheorem{lemma}{Lemma}
\newtheorem{definition}{Definition}

\definecolor{iccvblue}{rgb}{0.21,0.49,0.74}
\usepackage[pagebackref,breaklinks,colorlinks,allcolors=iccvblue]{hyperref}

\def\paperID{11437} 
\def\confName{ICCV}
\def\confYear{2025}

\title{LoRAverse: A Submodular Framework to Retrieve Diverse Adapters for Diffusion Models}

\author{Mert Sonmezer$^{*}$ \\
Middle East Technical University \\
{\tt\small mert.sonmezer@metu.edu.tr}
\and
Matthew Zheng \\
Virginia Tech \\
{\tt\small matthewz03@vt.edu}
\and
Pinar Yanardag \\
Virginia Tech \\
{\tt\small pinary@vt.edu}
}

\begin{document}
\twocolumn[{
\maketitle
\begin{center}
    \captionsetup{type=figure}
    \vspace{-1em}
    \newcommand{\imwidth}{\textwidth}

    \begin{tabular}{@{}c@{}}
        \parbox{\imwidth}{\includegraphics[width=\imwidth]{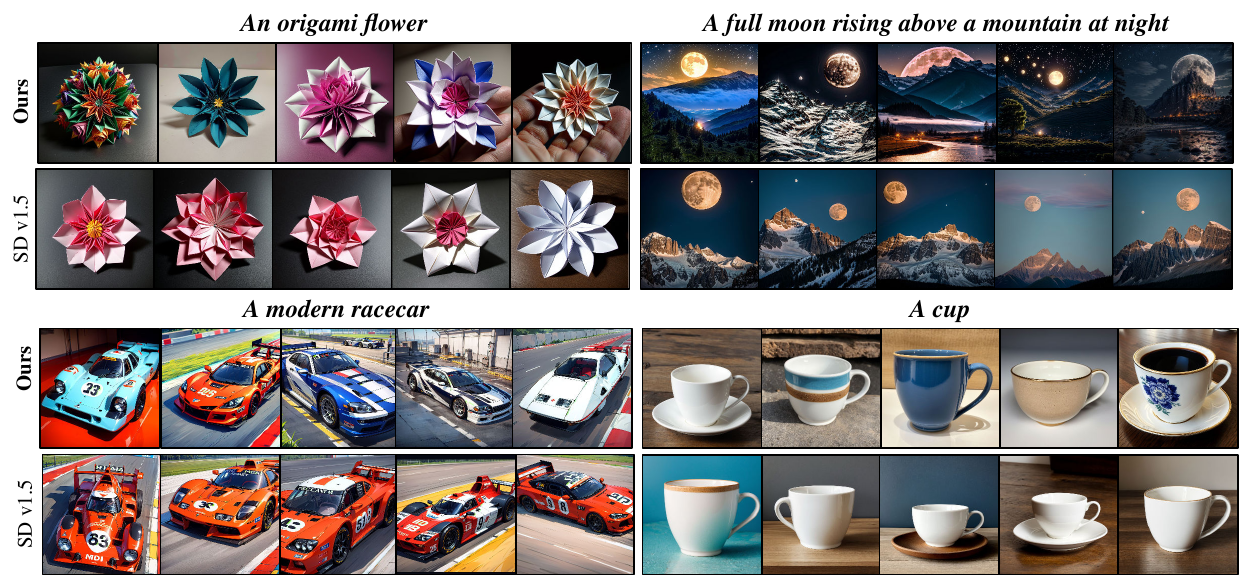}} \\
        \vspace{1em}
    \end{tabular}

    \vspace{-2.5em}
    \captionof{figure}{\textbf{Diverse Adapter Selection.} LoRAverse enhances image diversity by employing a submodular algorithm to select a diverse and representative set of LoRA adapters. This approach begins by clustering the adapters based on their semantic meanings. From these clusters, the algorithm selects models that not only maximize diversity but also maintain strong alignment with the user-provided prompt, ensuring both variety and relevance in the generated images.
    }
\end{center}
}]
\begingroup
\renewcommand\thefootnote{*}
\footnotetext{This work was conducted while the author was a research intern at Virginia Tech.}
\renewcommand\thefootnote{$\dagger$}
\footnotetext{Published in the Proceedings of the IEEE/CVF International Conference on Computer Vision (ICCV 2025), Honolulu, HI.}
\endgroup
\begin{abstract} 
Low-rank Adaptation (LoRA) models have revolutionized the personalization of pre-trained diffusion models by enabling fine-tuning through low-rank, factorized weight matrices specifically optimized for attention layers. These models facilitate the generation of highly customized content across a variety of objects, individuals, and artistic styles without the need for extensive retraining. Despite the availability of over 100K LoRA adapters on platforms like Civit.ai, users often face challenges in navigating, selecting, and effectively utilizing the most suitable adapters due to their sheer volume, diversity, and lack of structured organization. This paper addresses the problem of selecting the most relevant and diverse LoRA models from this vast database by framing the task as a combinatorial optimization problem and proposing a novel submodular framework. Our quantitative and qualitative experiments demonstrate that our method generates diverse outputs across a wide range of domains.
\end{abstract}

\section{Introduction}
\label{sec:intro}

LoRA \cite{ryu2023low} models have emerged as a powerful technique for model personalization. They can efficiently fine-tune pre-trained diffusion models without the need for extensive retraining or significant computational resources. They are designed to optimize low-rank, factorized weight matrices specifically for the attention layers and are typically used in conjunction with personalization methods such as DreamBooth \cite{ruiz2023dreambooth} to generate personalized content on specific objects, individuals, or artistic styles.  Since their introduction, LoRA models have gained widespread popularity among researchers, developers, and artists \cite{gandikota2023concept, guo2023animatediff}. Platforms like Civit.ai \cite{civitai_website}, which host over 100,000 LoRA models, have become major hubs for creators looking to extend the capabilities of pre-trained models to generate highly specific image styles, themes, or artistic directions. This flexibility allows users to target precise modifications without retraining the entire model, enabling fast adaptation and diverse creative outputs.  

Despite the vast potential, the sheer volume of available LoRA adapters makes it challenging for users to navigate and effectively use these models. Users often struggle to explore, evaluate, and combine different LoRA adapters to achieve desired outcomes. Without clear guidance or an efficient system for selecting the right combination, users are left with a trial-and-error approach that is both time-consuming and inefficient. Consider a user who wants to generate fantasy world concept designs with themes like "ancient ruins," "futuristic cities," and "enchanted forests" using LoRA adapters. Users familiar with platforms like Civit.ai  face an overwhelming choice among 100,000+ models, each varying in style, quality, and effectiveness. The challenge lies in selecting the best-fit adapter for their concept amid the massive range and diversity of LoRAs. Existing retrieval methods \cite{luo2024stylus} use top-$K$ ranking and LLM selection. While relevant, top-$K$ ranking often surfaces similar models, leading to redundant recommendations and overlapping features, and relying on LLMs to prune these further limits the variation and biases the input prompt.


In this paper, we explore the challenge of selecting the most relevant and diverse LoRA adapters from the vast database. We approach the task of finding suitable LoRA adapters, each tuned to generate highly specific styles and concepts, as a combinatorial optimization problem. Our objective is to identify a subset of LoRA adapters that are not only relevant but also diverse in the types of concepts they represent. This is crucial as many adapters may offer overlapping functionalities. Our objective function incorporates diversity by penalizing the selection of additional models from already-represented LoRA concepts. In order to achieve this, we cluster similar LoRA models, allowing our algorithm to prioritize unexplored clusters, thus enhancing the diversity of the selected subset. Our framework, LoRAverse, then ensures that selecting a LoRA from a cluster that has not yet been explored yields a higher gain. We formulate this task as a monotone submodular function maximization, for which there is a simple greedy algorithm guarantees that the solution obtained is $\epsilon$-close to the best possible solution \cite{lin2011class}.
\begin{figure*}[]
    \centering
    \includegraphics[width=\linewidth]{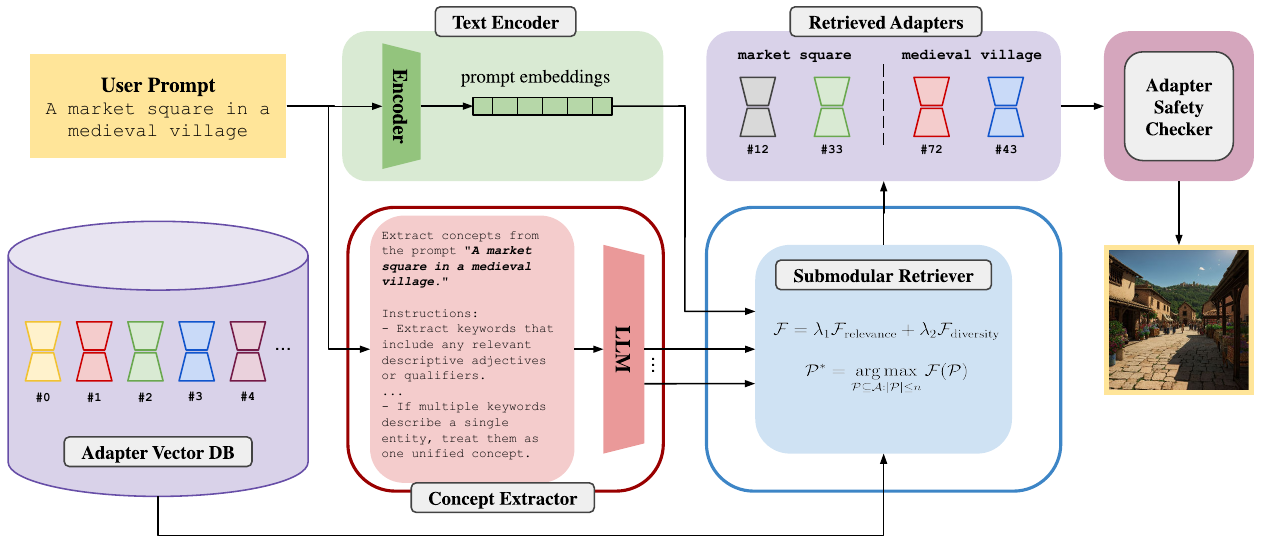}
    \caption[]{\textbf{Architecture of LoRAverse.} LoRAverse composed of two main modules: \textit{concept extractor} and \textit{submodular retriever}. The \textit{concept extractor} processes the user prompt to identify concepts (keywords). These concepts guide to the \textit{submodular retriever}, which selects a diverse and relevant subset of LoRA adapters by clustering similar adapters per concept and applying submodular optimization. Additionally, a safety-checking mechanism is integrated to filter out adapters containing offensive or inappropriate content. 
}\label{fig:architecture}
\end{figure*}

\section{Related Works}
\label{sec:related}

LoRA models provide an efficient solution for fine-tuning models on specific tasks with minimal parameter updates, achieving performance comparable to full fine-tuning while reducing computational and storage demands \cite{gal2022imageworthwordpersonalizing, ha2016hypernetworks, hu2021loralowrankadaptationlarge}. Their adaptability and efficiency have led to widespread adoption within open-source communities \cite{wolf2020huggingfacestransformersstateoftheartnatural}. In the image domain, these models excel at combining diverse attributes--such as styles, characters, poses, and actions--to generate high-quality images that align with user prompts \cite{liu2023unsupervisedcompositionalconceptsdiscovery, zhong2024multiloracompositionimagegeneration}. However, with the abundance of available adapters, selecting the most relevant ones for specific generative tasks remains a significant challenge \cite{Modelverse2023}. Retrieval-based methods like retrieval-augmented generation (RAG) address similar challenges by embedding text inputs into vectors and ranking results based on semantic similarity to the user prompt \cite{devlin2019bertpretrainingdeepbidirectional, gao2024retrievalaugmentedgenerationlargelanguage, lewis2020retrieval, lin2023vectorsearchopenaiembeddings, radford2021learning, reimers2019sentencebertsentenceembeddingsusing, luo2024stylus}. While effective, these methods often struggle to introduce diversity in the retrieved outputs. Submodular optimization, widely used for selecting diverse subsets from larger pools, provides an effective solution to this problem \cite{lin2011class}. In particular, submodular techniques model the trade-off between relevance and diversity in adapter selection, ensuring that both highly relevant adapters and diverse image styles are maintained. Clustering methods, such as those used in GAN-based models to group semantically similar elements \cite{collins2020editingstyleuncoveringlocal, pakhomov2021segmentationstyleunsupervisedsemantic, simsar2022fantasticstylechannelsthem}, further complement this by grouping similar adapters, enabling the selection of diverse yet relevant adapters for high-quality image generation. Recent diffusion‑based sampling strategies also tackle diversity, but at generation time rather than retrieval \cite{corso2023particleguidancenoniiddiverse}.
\section{Background}

Let $\mathcal{V}$ denote a set of elements $\mathcal{V} = \{v_1, \ldots v_n\}$ (the \textit{ground set}), and let $\mathcal{F}: 2^\mathcal{V} \to \mathbb{R}$ represent a function that assigns a real value to any subset $\mathcal{P} \subseteq \mathcal{V}$. The goal is then to select a small subset $\mathcal{P}^*\subseteq \mathcal{V}$ of size $|\mathcal{P}| \leq n$ that maximizes the function $\mathcal{F}(\mathcal{P})$, i.e., $\mathcal{P}^*\in\argmax_{\mathcal{P} \subseteq \mathcal{V},\, |\mathcal{P}|\leq n} \mathcal{F}(\mathcal{P})$. Solving this problem is intractable in general, but it has been shown that a greedy algorithm can be used to solve this equation almost optimally with an approximation factor of ($1-1/e$), under the condition that the function $\mathcal{F}$ is monotone, non-decreasing, and submodular \cite{sviridenko2004note}. Starting with an empty set $\mathcal{P}_0=\emptyset$, the algorithm iteratively selects the element $v_i$ that maximizes the marginal gain: $v_i\in \argmax_{v\in\mathcal{V}\setminus \mathcal{P}_{i-1}} \mathcal{F}(\mathcal{P}_{i-1}\cup\{v\}) - \mathcal{F}(\mathcal{P}_{i-1})$. After $n$ iterations, the resulting subset $\mathcal{P}^*$ satisfies the following inequality: $\mathcal{F}(\mathcal{P}^*)\geq (1-1/e)\mathcal{F}(\mathcal{P_{\text{opt}}})\approx 0.63\,\mathcal{F}(\mathcal{P}_{\text{opt}})$ where $\mathcal{P}_{\text{opt}}$ is the optimal subset. More formally, submodularity is defined as follows: 

\vspace{-8pt}
\begin{definition}
\label{def:submodularity}
The function $\mathcal{F}$ is called submodular if for all subsets $R\subseteq P \subseteq \mathcal{V}$ and any elements $v \in \mathcal{V} \setminus \mathcal{P}$, the following inequality holds: $\mathcal{F} (\mathcal{P} \cup \{ v\} ) - \mathcal{F}(\mathcal{P}) \leq \mathcal{F} (\mathcal{R} \cup \{ v \} ) - \mathcal{F}(\mathcal{R})$.
This form of submodularity directly satisfies the diminishing returns property:
the {\em value} of adding $v$ never becomes larger as the context becomes larger \cite{nemhauser1978analysis}.
\end{definition}
\newcommand{\qed}{\hfill \ensuremath{\Box}}

\section{Methodology}
\label{sec:methodology}

Selecting relevant LoRA models is challenging, especially compared to text-based document retrieval methods (\S~\ref{sec:related}). User prompts often contain multiple nuanced concepts, making it difficult to generate high-quality, diverse images. The challenge lies not only in retrieving LoRA models that fit the entire prompt but also in correctly associating them with specific concepts while maintaining diversity. The large number of similar models in the database further complicates achieving diversity and quality. Additionally, combining multiple LoRAs can degrade image quality. To address these issues, we propose LoRAverse (Fig. \ref{fig:architecture}), a novel framework that frames the task as a combinatorial optimization problem to effectively address these challenges.

\subsection{Concept Extractor}
\label{subsec:concept-extractor}
The \textit{concept extractor} is a concise, single-step pipeline built on a large language model (LLM), designed to divide a user prompt into distinct, non-overlapping chunks; full prompt details are provided in Supplementary Material (SM) \S~\ref{sup:concept-extractor-details}.. This segmentation process facilitates the extraction of nuanced concepts from the prompt, enabling the retrieval of adapters that are conceptually aligned with the overall prompt. The system extracts two distinct concepts, for example, ``\textit{British shorthair cat}" and ``\textit{cherry blossom garden}," for the provided prompt ``\textit{a British shorthair cat playing in a cherry blossom garden.}" The \textit{concept extractor} is specifically tasked with extracting such key concepts from the user's prompt. Mathematically, the \textit{concept extractor} $\mathcal{C}$ processes a given prompt $s$, extracting a set of concepts $\mathcal{T}(s) = \{t_1, t_2, \dots, t_n\}$. This process is defined as:

\begin{equation}
\label{eq:concept-extractor}
\mathcal{C}(s) = \{t_i | t_i \in \mathcal{T}(s), t_i \subseteq s\}
\end{equation}

\subsection{Submodular Retriever}
\label{subsec:retriever}

In our work, we treat user prompts as combinations of distinct concepts (\S~\ref{subsec:concept-extractor}) and aim to retrieve the top-$K$ most relevant and diverse LoRA models based on these concepts. A simple approach would be retrieving adapters using cosine similarity between the full prompt embedding and adapter embeddings. While this ensures relevance, it often results in redundancy, as retrieved adapters show only minor variations, limiting diversity (see Fig. \ref{fig:alternative-retrieval-method}). On the other hand, selecting adapters based solely on individual concept embeddings can introduce adapters relevant to specific concepts but misaligned with the overall prompt meaning (see Fig. \ref{fig:varying-relevance-trade-off}). To address these issues, our \textit{submodular retriever} module maximizes a monotone submodular function, balancing relevance and diversity. Using a greedy algorithm, we find an almost-optimal solution \cite{lin2011class}, ensuring the selected adapters are both relevant and diverse. Formally, let $\mathcal{A}$ be the set of available LoRA models, and we select a diverse, representative subset $\mathcal{P} \subseteq \mathcal{A}$, where relevance is based on the similarity between the prompt embedding and LoRA model embeddings from a Vision Language Model (VLM). We define the relevance objective as:

\begin{equation}
\label{eq:relevance}
\mathcal{F}_{\text{relevance}}(\mathcal{P}) = \sum_{a_i \in \mathcal{P}} \mathcal{F}_{\text{sim}}(\phi(a_i), \phi(s))
\end{equation}

\noindent where $\phi(s)$ and $\phi(a_i)$ denote the embeddings of the user prompt $s$ and the metadata of the LoRA model $a_i$ respectively. This equation quantifies how well the set $\mathcal{P}$ aligns with the user prompt. The function $\mathcal{F}_{\text{sim}}$ measures similarity, which can be computed using cosine similarity.

However, this relevance-only function neglects diversity, often favoring multiple models of the same style (e.g., photorealistic) and overlooking others \cite{civitai_website}. To address this, a common approach applies a diversity regularization to the objective function \cite{lin2011class, simsar2022fantasticstylechannelsthem}, rewarding items selected from different clusters such that:

\begin{equation}
\label{eq:diversity}
\mathcal{F}_{\text{diversity}}(\mathcal{P}) = \sum_{k=1}^K  \left( \log  \left(1+ \sum_{a_i \in \mathcal{C}_k \cap \mathcal{P}}   \mathcal{F}_{\text{reward}}({\phi(a_i)}) \right) \right) 
\end{equation}

\noindent Here, the set of LoRA models $\mathcal{A}$ is divided into $K$ disjoint clusters $\mathcal{C}_k$, where  $k=1, \ldots, K$ and  $\bigcup_k \mathcal{C}_k = \mathcal{A}$. Each LoRA model $a_i$ has a reward $\mathcal{F}_{\text{reward}}({\phi(a_i)}) \geq 0$, indicating its importance in adding $a_i$ to the empty set.  This reward is computed as the cosine similarity between the embedding $\phi(a_i)$ of the LoRA adapter and the concept extracted from the user prompt.

The clustering objective ensures selecting models from unexplored clusters yields higher gains than from already-covered ones, balancing prompt relevance and diversity. To avoid the inefficiency of clustering the entire LoRA set, we first retrieve a subset relevant to the prompt using cosine similarity, then apply HDBSCAN \cite{Malzer_2020} on their textual embeddings via BERTopic \cite{grootendorst2022bertopic} (see SM \S~\ref{sup:clustering} for details). This groups LoRA models semantically, ensuring diverse and relevant selection. Then, the overall objective combines both:

\begin{equation}
\mathcal{F}(\mathcal{P}) = \lambda_1 \mathcal{F}_{\text{relevance}}(\mathcal{P}) + \lambda_2 \mathcal{F}_{\text{diversity}}(\mathcal{P})
\label{eq:objective}
\end{equation}

\noindent Here, $\lambda_1\geq0$ and $\lambda_2\geq0$ control the trade-off between relevance and diversity. To obtain a subset of LoRA models, we aim to maximize:

\begin{equation}
\label{eq:argmax}
\mathcal{P}^* = \argmax_{\mathcal{P} \subseteq \mathcal{A},\,|\mathcal{P}|\leq n} \mathcal{F}(\mathcal{P})
\end{equation}

\noindent where $n$ represents the cardinality constraint, denoting the maximum number of elements allowed in the selected subset $\mathcal{P}^*$. This approach ensures semantic alignment with the prompt while maintaining diversity and avoiding redundancy. 

Finding the optimal subset is computationally infeasible, but a greedy algorithm can approximate the solution within a factor of $1-1/e$ for monotone, non-negative, submodular functions \cite{nemhauser1978analysis}. In practice, these algorithms often surpass this theoretical bound, proving both efficient and reliable \cite{lin2011class}. The \textit{submodular retriever} brings this approach together by selecting relevant and diverse LoRA models for each concept while aligning with the full prompt. It clusters similar models and applies submodular function maximization. Formally, the \textit{submodular retriever} $\mathcal{R}$ takes a set of concepts $\mathcal{T}(s)$ from the \textit{concept extractor} and retrieves the adapter set $\mathcal{P}^*$:

\begin{equation}
\label{eq:retriever}
\mathcal{R}(\mathcal{T}(s)) = \{a_i| a_i \in \mathcal{P}^*_{t_i}, \mathcal{P}^*_{t_i} \subseteq \mathcal{A}, t_i\in\mathcal{T}(s)\}
\end{equation}

\noindent where each $a_i$ represents an adapter in the selected subset $\mathcal{P}^*_{t_i}$ for the concept $t_i$. The submodular function is defined and computed separately for each concept (see SM \S~\ref{sup:proof} for the complete proof). Finally, LoRAverse generates diverse image sets by randomly selecting and linearly combining the LoRA models retrieved by the \textit{submodular retriever} for each concept.

\begin{figure*}[]
    \centering
    \includegraphics[width=0.96\linewidth]{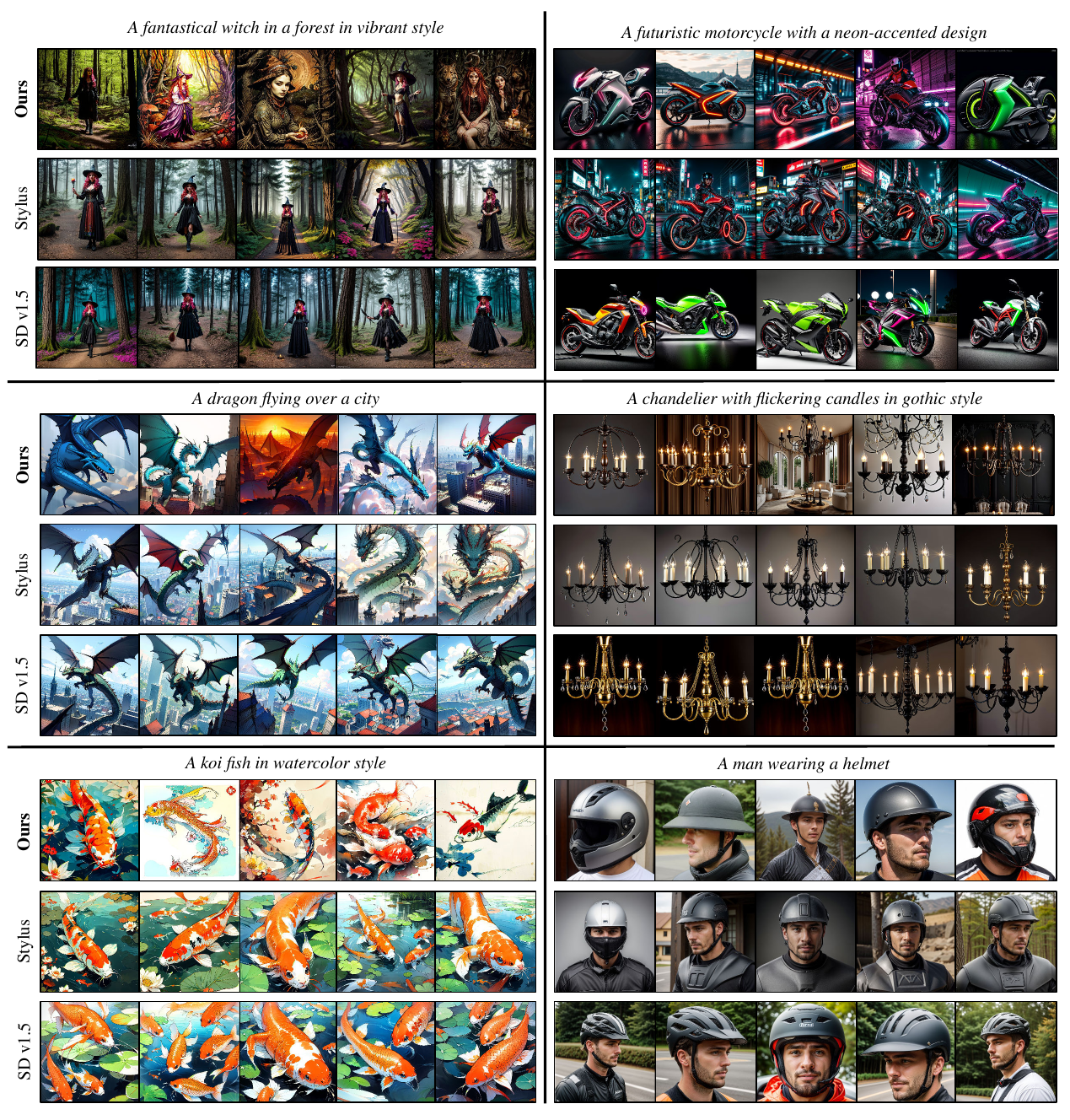}
    \caption[]{\textbf{Qualitative Comparison.} LoRAverse demonstrates a higher diversity compared to image sets generated by Stylus and SD v1.5.
}\label{fig:qualitative-comparison}
\end{figure*}

\section{Experiments}
\label{sec:experiments}


We compare LoRAverse with two baseline models: Stylus \cite{luo2024stylus} and Stable-Diffusion-v1.5 (SD v1.5) \cite{rombach2022high}, using the two of the most prominent model checkpoints \texttt{Realistic-Vision-v6} and \texttt{Counterfeit-v3}. We used StylusDocsv2 \cite{luo2024stylus} database, which includes around 75,000 LoRA models from platforms like CivitAI \cite{civitai_website} and Hugging Face \cite{hf_website}.  Images are generated via the Stable Diffusion Web UI \cite{AUTOMATIC1111_Stable_Diffusion_Web_2022} with 35 denoising steps and the DPM Solver++ scheduler \cite{lu2023dpmsolverfastsolverguided}. During the experiments, we observed occasional style drift caused by certain LoRA models, which we mitigated by appending a debias prompt\footnote{The debias prompt is "realistic, high quality" for \texttt{Realistic-Vision-v6} and "anime style, high quality" for \texttt{Counterfeit-v3}} to the user input, ensuring style consistency—similar to Stylus. Additionally, to maintain the ethical integrity of LoRAverse, we implemented an adapter safety checker using \texttt{gpt-4o} (see SM \S~\ref{sup:safety-checker-details} for the full prompt) to filter out adapters containing inappropriate or explicit content. This ensures that the retrieved adapters align with professional expectations and ethical standards. In our submodular algorithm, we set the relevance and diversity trade-off parameters, $\lambda_1$ and $\lambda_2$, to 7.0 and 1.0, respectively, balancing both factors in the selection process.

\begin{table}[t]
\centering
\begin{tabular}{lccc}
\hline
& \textbf{LoRAverse} & Stylus & SD v1.5 \\ 
\hline
CLIP ($\uparrow$) & 25.07 (\textcolor{red}{-3.1\%}) & 25.41 (\textcolor{red}{-1.8\%}) & \textbf{25.88} \\
TCE  ($\uparrow$) & \textbf{22.63} (\textcolor{green}{16.5\%}) & 20.30 (\textcolor{green}{4.5\%}) & 19.43 \\
TIE  ($\uparrow$) & \textbf{40.06} (\textcolor{green}{5.1\%}) & 38.53 (\textcolor{green}{1.1\%}) & 38.12 \\
I2I  ($\downarrow$) & \textbf{0.784} (\textcolor{green}{-7.3\%}) & 0.825 (\textcolor{green}{-2.5\%}) & 0.846 \\
\hline
US-F ($\uparrow$) & \textbf{3.78} (\textcolor{green}{4.7\%}) & 3.71 (\textcolor{green}{2.8\%}) & 3.61 \\
US-D ($\uparrow$) & \textbf{3.92} (\textcolor{green}{39.5\%}) & 2.82 (\textcolor{green}{0.4\%}) & 2.81 \\
US-P ($\uparrow$) & \textbf{47.55\%} & 29.90\% & 22.55\% \\
\hline
\end{tabular}
\caption{\textbf{Quantitative Comparison (CFG=7).} LoRAverse enhances the diversity of image sets across various metrics while maintaining comparable text-image alignment. The user study reports which method produced preferred outputs (US-P) by participants, and average rating of faithfulness (US-F) and diversity (US-D) of outputs on a scale of 1 to 5.}
\label{tab:quantitave-comparison}
\end{table}

\subsection{Qualitative Experiments}
\label{subsec:qualitative}

We compare LoRAverse qualitatively with the baseline approaches in Fig. \ref{fig:qualitative-comparison}. Each comparison includes 15 images generated by our method, Stylus \cite{luo2024stylus}, and SD v1.5 \cite{rombach2022high}, showing that our approach consistently produces the most diverse image sets. In contrast, the baseline methods often struggle to retrieve adapters that produce images with a wide range of diversity in aspects such as style, background, or primary content. While Stylus provides greater diversity than SD v1.5, its reliance on adapter retrieval based solely on cosine similarity leads to redundancy among the selected adapters (\S~\ref{subsec:ablation}). Furthermore, its use of an LLM called \textit{composer} \cite{luo2024stylus} for selection introduces an inherent bias toward the input prompt, limiting the diversity of its outputs. Consequently, it cannot surpass a certain threshold in generating varied images. Our framework overcomes these limitations with a submodular-based adapter selection process, driven entirely by an algorithmic framework. For example, given the prompt "\textit{A fantastical witch in a forest in vibrant style}," other methods struggle to generate witches with distinct clothing or diverse settings. In contrast, our method effectively captures and enhances this diversity, delivering a richer and more varied visual representation. Additional qualitative results and further examples are provided in SM \S~\ref{sup:qualitative-results}.

\begin{figure}[]
    \centering
    \includegraphics[width=\linewidth]{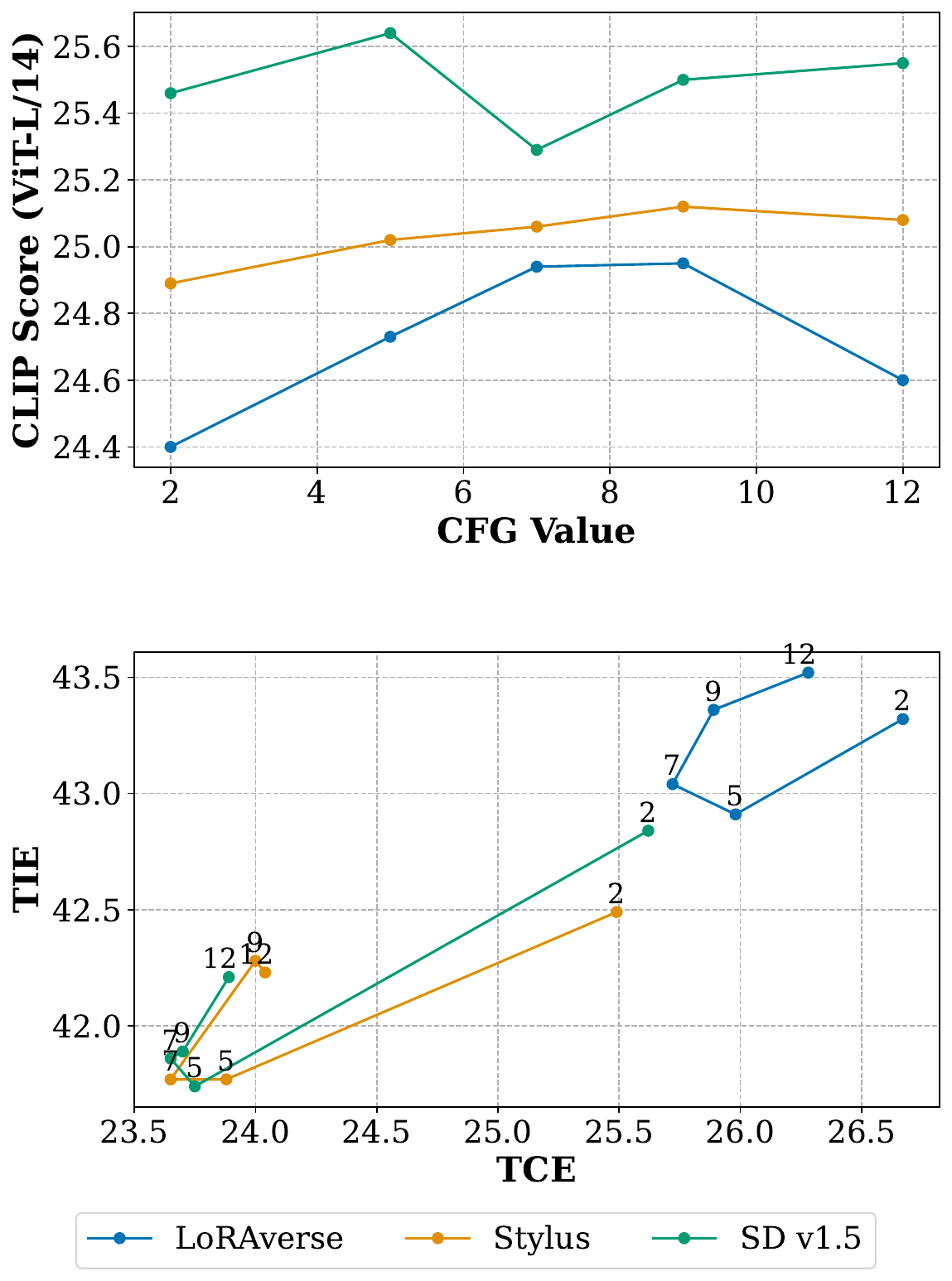}
    \caption[]{\textbf{Impact of CFG values.} The top shows CLIP score variations across CFG values, with LoRAverse remaining competitive despite a slight decrease. The bottom highlights the TCE-TIE trade-off with varying CFG values, where LoRAverse consistently achieves higher diversity, pushing the Pareto frontier toward greater semantic and visual diversity.
}\label{fig:varying-cfg}
\end{figure}

\subsection{Quantitative Experiments}
\label{subsec:quantitative}

\noindent\textbf{Image Diversity.} We evaluate LoRAverse quantitatively using four benchmarks: CLIP (ViT-L/14) \cite{CLIPScore}, Truncated CLIP Entropy (TCE) \cite{ibarrola2024measuringdiversitycocreativeimage}, Truncated Inception Entropy (TIE) \cite{ibarrola2024measuringdiversitycocreativeimage}, and pairwise image similarity (I2I). CLIP measures text-image alignment, while TCE and TIE assess diversity--TCE captures semantic diversity in CLIP's text-image embedding space, and TIE quantifies visual diversity using InceptionV3 \cite{szegedy2015rethinkinginceptionarchitecturecomputer}. Additionally, I2I measures pairwise similarity using image-to-image CLIP score.

Our evaluation includes 15,000 images--5,000 per method and 20 per prompt--from MS-COCO \cite{lin2015microsoftcococommonobjects} and PartiPrompts \cite{yu2022parti} using the \texttt{Realistic-Vision-v6} checkpoint. We retrieve top-200 LoRA models per concept and selected top-8 of them using our method. As shown in Table \ref{tab:quantitave-comparison}, LoRAverse significantly enhances diversity while maintaining strong text-image alignment. Despite a 3.1\% decrease in CLIP score, it compensates with a 16.5\% increase in TCE and a 5.1\% increase in TIE, demonstrating superior diversity. Additionally, its outputs are 7.3\% more dissimilar than those of baseline methods, as reflected in I2I scores. Further quantitative experiments can be found in SM \S~\ref{sup:quantitative-results} and \ref{sup:cluster-concept-ablation}.

LoRAverse preserves diversity across varying Classifier-Free Guidance (CFG) values (Fig. \ref{fig:varying-cfg}) and shifts the Pareto curve toward greater semantic and visual diversity, demonstrating its effectiveness. This confirms that our submodular-based approach outperforms LLM-based methods, like Stylus, in promoting diversity, making LoRAverse more versatile. The slight drop in CLIP score is expected, as increased diversity naturally broadens the output distribution, leading to greater variation in image composition. CLIP favors strict prompt adherence over generative diversity due to its training on large-scale web image-text pairs, which reinforce common associations over novel representations \cite{alabdulmohsin2024clipbiasusefulbalancing, shao2023investigatinglimitationclipmodels}. Thus, the lower CLIP score reflects LoRAverse's improved diversity rather than a decline in text-image alignment. \\

\begin{figure}[]
    \centering
    \includegraphics[width=\linewidth]{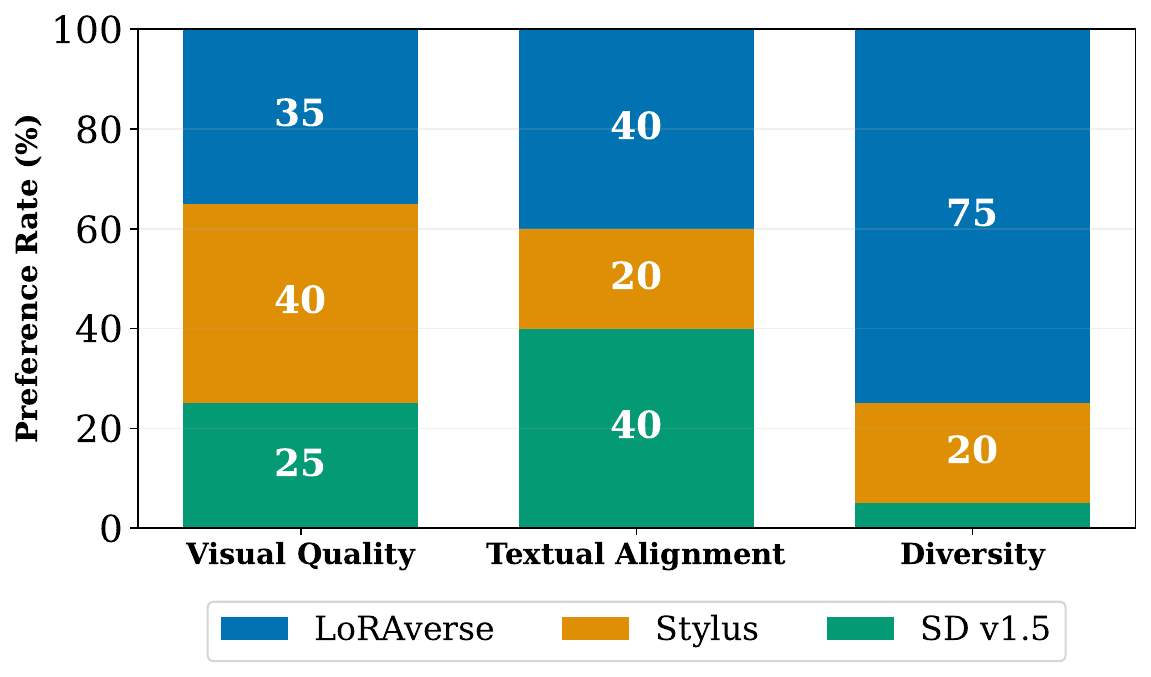}
    \caption[]{\textbf{Preference Rate of VLM-as-a-Judge.} Preference rates of LoRAverse, Stylus, and SD v1.5 across visual quality, textual alignment, and diversity, as evaluated by \texttt{gpt-4o}. LoRAverse achieves a clear advantage in diversity while maintaining competitive visual quality and textual alignment.
}\label{fig:vlm-as-a-judge}
\end{figure}

\begin{figure}[]
    \centering
    \includegraphics[width=\linewidth]{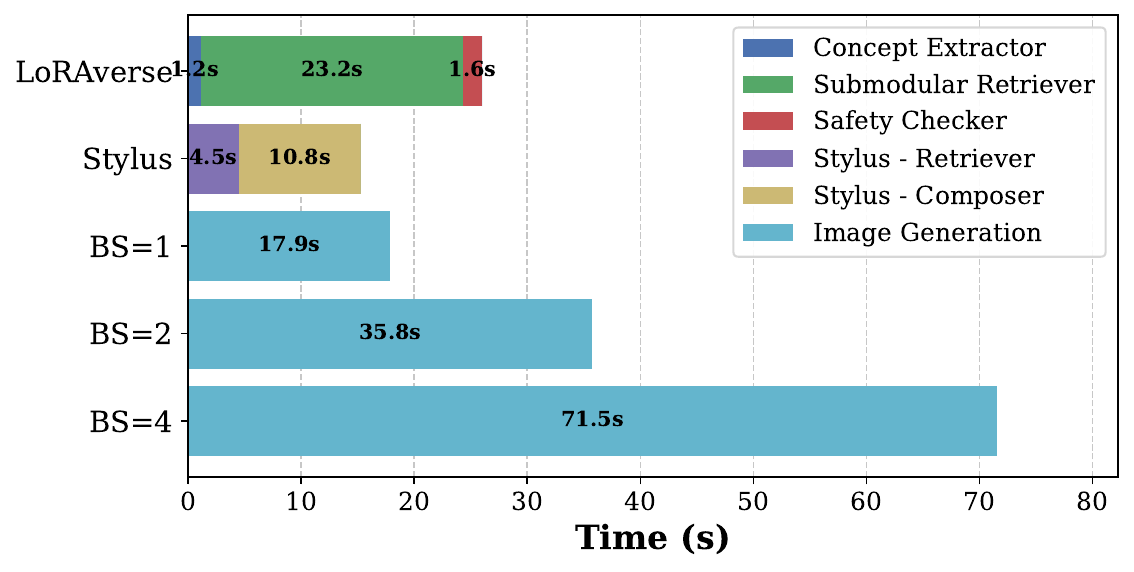}
    \caption[]{\textbf{Inference Time Analysis.} Comparison of LoRAverse's inference time overheads for single-concept prompts with Stylus's and SD v1.5's inference time for varying batch size (BS). The increased overhead in LoRAverse is primarily due to the time spent on clustering adapters, but this overhead decreases proportionally as the batch size increases.
}\label{fig:inference-time-analysis}
\end{figure}

\noindent\textbf{VLM-as-a-Judge.} We adopt the VLM-as-a-Judge approach to evaluate the image quality, textual alignment, and diversity of LoRAverse, Stylus, and SD v1.5, generating five images per prompt across 300 images with positions randomly shuffled to minimize bias \cite{zhong2024multiloracompositionimagegeneration}. We ask \texttt{gpt-4o} to rate images from 0 (poor) to 2 (high) for quality, alignment, and diversity. As shown in Fig. \ref{fig:vlm-as-a-judge}, LoRAverse demonstrates a clear advantage in diversity, significantly outperforming both Stylus and SD v1.5. It also maintains strong performance in textual alignment and visual quality, closely matching or surpassing the baselines. These results highlight LoRAverse's ability to balance diversity and quality. Full prompt details and settings are provided in SM \S~\ref{sup:vlm-as-a-judge}. \\

\noindent\textbf{User Study.} To showcase our method's efficacy with human evaluators, we conduct a user study on 120 images that presents questions encompassing preference of generation, faithfulness to text prompt, and image diversity to 51 anonymous participants. For each question, users are asked to compare results from LoRAverse, Stylus, and SD v1.5. In Table \ref{tab:quantitave-comparison}, users demonstrated higher preference for LoRAverse over Stylus and SD v1.5. When assessing faithfulness, LoRAverse is competitive when users rated images, but when comparing diversity, users displayed strong preference over other methods, buttressing LoRAverse's ability to promote diverse image generation. Further details and results can be found in SM \S~\ref{sup:user-study}. \\

\noindent\textbf{Inference Time Analysis.} We analyze the inference time breakdown of LoRAverse's components, noting that SD v1.5 handles image generation independently. LoRAverse adds an average of 26 s to SD v1.5's generation time, with the \textit{submodular retriever} contributing 23.2 s due mainly to the clustering of retrieved adapters (see Fig. \ref{fig:inference-time-analysis}). Unlike Stylus, LoRAverse performs all adapter selection process locally for each individual concept without relying on external LLM APIs, which adds to this overhead. Furthermore, since LoRAverse clusters adapters for each individual concept, the time spent in the \textit{submodular retriever} increases by 10.54 s for each additional concept in the prompt. This overhead remains constant regardless of batch size, as each component is executed just once per inference. Additionally, increasing the number of clusters per concept has only a minor effect: expanding from 1 to 50 clusters raises the adapter selection time from 0.07 s to 1.86 s. To wrap up, for batch processing, the overhead of LoRAverse decreases as the batch size grows.

\begin{table}[t]
\centering
\begin{tabular}{lcc}
\hline
& \textbf{Ours} & Cosine Similarity \\ 
\hline
CLIP ($\uparrow$) & 24.50 (\textcolor{red}{-0.7\%}) & \textbf{24.67} \\
TCE  ($\uparrow$) & \textbf{23.97} (\textcolor{green}{2.1\%}) & 23.47 \\
TIE  ($\uparrow$) & \textbf{42.43} (\textcolor{green}{1.0\%}) & 41.99 \\
I2I  ($\downarrow$) & \textbf{0.762} (\textcolor{green}{-2.4\%}) & 0.781 \\
\hline
\end{tabular}
\caption{\textbf{Quantitative Comparison of Retrieval Algorithms.} Our submodular-based retrieval algorithm raises the diversity scores along with a slight decrease in the CLIP score, comparing to pure cosine similarity-based retrieval, due to its relevance-focus behavior.}
\label{tab:algo-comparison}
\end{table}

\begin{figure*}[]
    \centering
    \includegraphics[width=\linewidth]{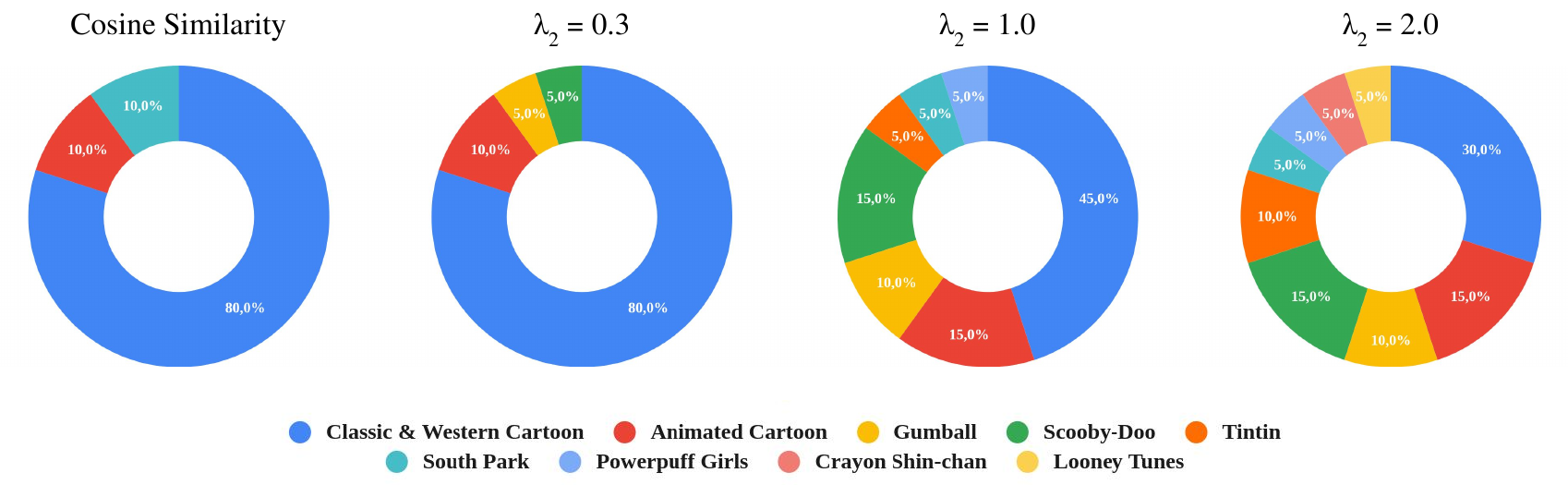}
    \caption[]{\textbf{Cluster Diversity.} Top-20 adapters for the concept \textit{cartoon character} from the prompt \textit{A cartoon character standing in a colorful fantasy world}, using cosine similarity and varying diversity trade-off ($\lambda_2$). Our submodular-based retrieval ensures diversity by selecting adapters from different clusters based on $\lambda_2$, while cosine similarity tends to retrieve adapters from the same clusters.
}\label{fig:cluster-diversity}
\end{figure*}

\begin{figure}[]
    \centering
    \includegraphics[width=\linewidth]{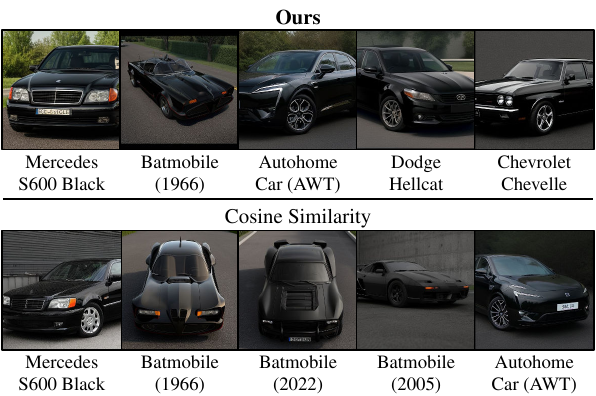}
    \caption[]{\textbf{Qualitative Comparison of Retrieval Algorithms.} Images generated from the prompt \textit{a black car} using the top-5 adapters retrieved by our submodular-based algorithm and cosine similarity. Our method selects diverse adapters from different clusters, while cosine similarity retrieves adapters with overlapping features, causing redundancy.
}\label{fig:alternative-retrieval-method}
\end{figure}

\begin{figure}[]
    \centering
    \includegraphics[width=0.9\linewidth]{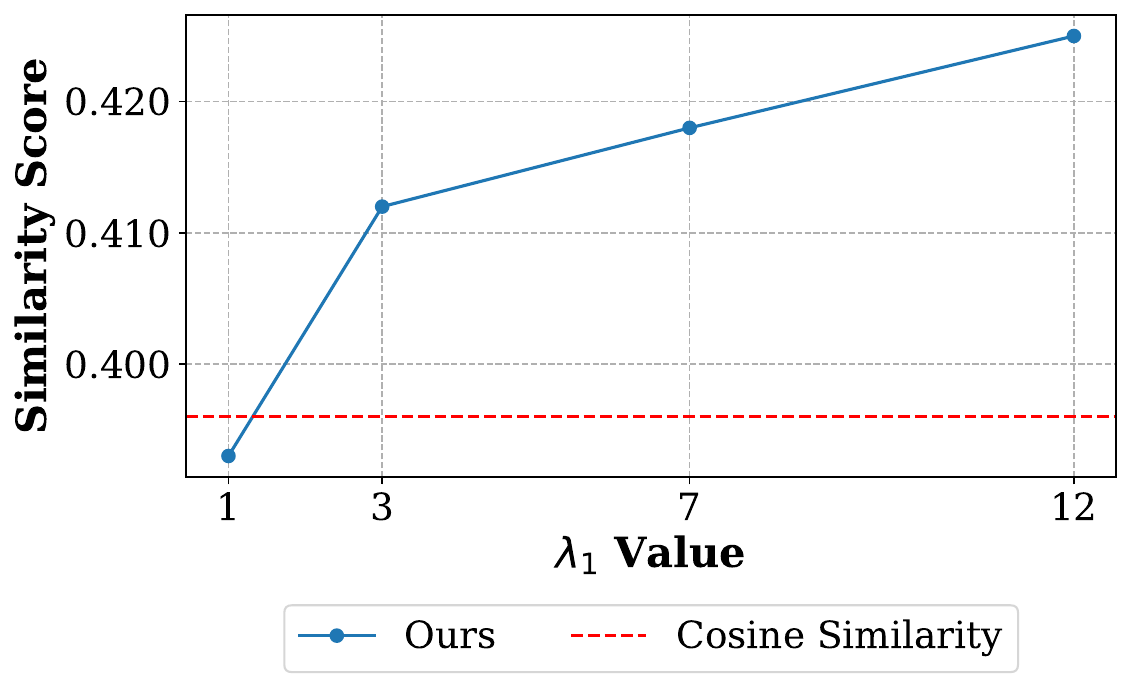}
    \caption[]{\textbf{Varying Relevance Trade-off ($\lambda_1$).} Cosine similarity scores between the top-20 adapters' descriptions, retrieved for the concept \textit{cartoon character}, and the user prompt \textit{A cartoon character standing in a colorful fantasy world}, evaluated with varying $\lambda_1$. The red line shows the similarity score of adapters retrieved using only cosine similarity.
}\label{fig:varying-relevance-trade-off}
\end{figure}

\subsection{Ablation Studies}
\label{subsec:ablation}

\noindent\textbf{Alternative Retrieval Methods.}  We compared our submodular-based retrieval algorithm with the widely used cosine similarity-based method. To create Table \ref{tab:algo-comparison}, we used 520 images in the same experimental setup in \S~\ref{sec:experiments}. As shown in both Fig. \ref{fig:alternative-retrieval-method} and Table \ref{tab:algo-comparison}, our approach optimizes for diversity, effectively retrieving LoRA models from different clusters. In contrast, retrieval based solely on cosine similarity between the adapters and full user prompt embeddings struggles to ensure diversity. For example, among the top-5 retrieved LoRAs (Fig. \ref{fig:alternative-retrieval-method}), three Batmobile adapters were selected, resulting in lack of variety. Our submodular approach, by selecting adapters across clusters, ensures more diverse image sets. Fig. \ref{fig:cluster-diversity} further shows that cosine similarity retrieves adapters from the same cluster, while our method distributes selections more evenly. \\

\noindent\textbf{Impact of Relevance and Diversity Trade-offs.} We analyze the impact of two key hyperparameters: relevance ($\lambda_1$) and diversity ($\lambda_2$), which control the balance between prompt relevance and the diversity of retrieved adapters. In Fig.~\ref{fig:varying-relevance-trade-off}, we examine how the relevance trade-off $\lambda_1$ affects the retrieval process using two different methods. The first method retrieves adapters purely based on cosine similarity between the concept and adapter embeddings. The second method applies our submodular-based retrieval algorithm while varying $\lambda_1$. The results show that as $\lambda_1$ increases, the retrieved adapters become more aligned with the full user prompt, rather than just the extracted concept, leading to higher overall prompt similarity. On the other hand, Fig. \ref{fig:cluster-diversity} demonstrates how $\lambda_2$ influences the distribution of adapters across clusters, with higher $\lambda_2$ encouraging more diverse selections. By adjusting these trade-offs, our method retrieves adapters are both relevant and diverse.

\section{Conclusion}
\label{sec:conclusion}

We present LoRAverse, a novel framework that autonomously selects LoRA models to generate diverse yet prompt-aligned image sets. Our two-stage architecture first decomposes user prompts into key concepts and then retrieves adapters that are both diverse and contextually relevant. Extensive evaluations  show that LoRAverse significantly outperforms existing methods in promoting diversity, offering a robust  solution for enhancing creative image generation while ensuring adaptability to various user prompts.
{
    \small
    \bibliographystyle{ieeenat_fullname}
    \bibliography{main}
}

%

\clearpage
\setcounter{page}{1}
\maketitlesupplementary

\section{Limitations}
\label{sec:limitations}

Although LoRAverse is able to select meaningful adapters for each extracted concept,  possible errors during clustering of adapters prior to applying submodular selection may lead to over-retrieval of similar adapters or neglect of relevant ones. If alike adapters were placed in the different clusters or dissimilar adapters are clustered together, rendered diversity could be limited. Moreover, retrieved LoRA models, which could inadvertently amplify societal biases present in training data. Therefore, we advocate for the implementation of safeguards to mitigate both bias and misuse risks.

\section{Proof of Submodularity}
\label{sup:proof}

To justify the use of a greedy algorithm for optimizing Eq. \ref{eq:objective}, we formally prove that the combined objective function $\mathcal{F}(\mathcal{P})$ is submodular. We begin by analyzing the components $\mathcal{F}_{\text{rel}}(\mathcal{P})$\footnote{$\mathcal{F}_{\text{rel}}$ is the abbreviated form of $\mathcal{F}_{\text{relevance}}$.} and $\mathcal{F}_{\text{div}}(\mathcal{P})$\footnote{$\mathcal{F}_{\text{div}}$ is the abbreviated form of $\mathcal{F}_{\text{diversity}}$.} separately.

\begin{lemma}
\label{lemma:relevance}
$\mathcal{F}_{\text{rel}}(\mathcal{P})$, formulated in Eq. \ref{eq:relevance}, is a modular function and therefore submodular.

\noindent Proof. A function is modular if it can be expressed as a linear sum over its elements, i.e., $\mathcal{F}(\mathcal{P}) = \sum_{v\in\mathcal{P}} w(v)$, for some function $w: \mathcal{V}\rightarrow \mathbb{R}$. Here, $\mathcal{F}_{\text{rel}}(\mathcal{P})$ is modular with $w(a_i) = \mathcal{F}_{\text{sim}}(\phi(a_i), \phi(s))$. For any $\mathcal{P}\subseteq \mathcal{V}$ and $v\notin \mathcal{P}$, the marginal gain of adding $v$ is:

\begin{equation}
    \mathcal{F}_{\text{rel}}(\mathcal{P}\cup \{v\})-\mathcal{F}_{\text{rel}}(\mathcal{P})=\mathcal{F}_{\text{sim}}(\phi(a_i), \phi(s))
\end{equation}

\noindent This marginal gain is independent of $\mathcal{P}$, satisfying the submodularity condition with equality:

\begin{equation}
    \mathcal{F}_{\text{rel}}(\mathcal{P}\cup \{v\})-\mathcal{F}_{\text{rel}}(\mathcal{P})= 
    \mathcal{F}_{\text{rel}}(\mathcal{R}\cup \{v\})-\mathcal{F}_{\text{rel}}(\mathcal{R})    
\end{equation}

\noindent for all $R\subseteq P$. Thus, $\mathcal{F}_{\text{rel}}(\mathcal{P})$ is modular and therefore submodular. \qed
\end{lemma}

\begin{lemma}
\label{lemma:diversity}
$\mathcal{F}_{\text{div}}(\mathcal{P})$, formulated in Eq. \ref{eq:diversity}, is submodular.

\noindent Proof. Suppose $v\in \mathcal{C}_k$ for some cluster $k$. Since the clusters $\{\mathcal{C_\text{1}, \dots, \mathcal{C}_\text{K}}\}$ are disjoint, adding $v$ only affects the term for $\mathcal{C}_k$ in the sum. For all other clusters $j \not= k$, the marginal gain is zero. Thus, the inequality reduces to:

\begin{equation}
\begin{split}
    \log(1+\mathcal{S_\mathcal{P}}&+\mathcal{F}_\text{rew}(v)) - \log(1+\mathcal{S_\mathcal{P}}) \leq  \\&\log(1+\mathcal{S_\mathcal{R}}+\mathcal{F}_\text{rew}(v)) - \log(1+\mathcal{S_\mathcal{R}})
\end{split}
\end{equation}

\noindent where $\mathcal{S_\mathcal{P}}=\sum_{a_i\in\mathcal{C}_k\cap\mathcal{P}}\mathcal{F}_{\text{rew}}(a_i)$\footnote{$\mathcal{F}_\text{rew}$ is the abbreviated form of $\mathcal{F}_\text{reward}$} and $\mathcal{S_\mathcal{R}}=\sum_{a_i\in\mathcal{C}_k\cap\mathcal{R}}\mathcal{F}_{\text{rew}}(a_i)$. The function $g(x)=\log(1+x)$ is concave, so its derivative $g'(x)=1/(1+x)$ is decreasing. By the Mean Value Theorem, for some $\xi\in[S_\mathcal{R}, S_\mathcal{R}+\mathcal{F}_{\text{rew}}(v)]$ and $\zeta\in[S_\mathcal{R}, S_\mathcal{R}+\mathcal{F}_{\text{rew}}(v)]$:

\begin{equation}
\begin{split}
     &\log(1+\mathcal{S_\mathcal{R}}+\mathcal{F}_\text{rew}(v)) - \log(1+\mathcal{S_\mathcal{R}}) = \frac{\mathcal{F_\text{rew}}(v)}{1+\xi} \\
     &\log(1+\mathcal{S_\mathcal{P}}+\mathcal{F}_\text{rew}(v)) - \log(1+\mathcal{S_\mathcal{P}}) = \frac{\mathcal{F_\text{rew}}(v)}{1+\zeta} 
\end{split}
\end{equation}

\noindent Since $\mathcal{R}\subseteq \mathcal{P}$, we have $S_\mathcal{R} \leq S_\mathcal{P}$, which implies $\xi \leq \zeta$. Therefore:

\begin{equation}
    \frac{\mathcal{F_\text{rew}}(v)}{1+\zeta} \leq \frac{\mathcal{F_\text{rew}}(v)}{1+\xi}
\end{equation}

\noindent This establishes the inequality. Since the inequality holds for every cluster $\mathcal{C}_k$, summing over all $k$ preserves submodularity. Thus, $\mathcal{F}_\text{div}(\mathcal{P})$ is submodular. \qed
\end{lemma}

\begin{lemma}
\label{lemma:objective}
$\mathcal{F}_(\mathcal{P})$, formulated in Eq. \ref{eq:objective}, is submodular.

\noindent Proof. Since $\mathcal{F}(\mathcal{P})$ is a non-negative linear combination of submodular functions, it is itself submodular. \qed
\end{lemma}

\section{Additional Qualitative Results}
\label{sup:qualitative-results}

\begin{figure*}[]
    \centering
    \includegraphics[width=0.88\linewidth]{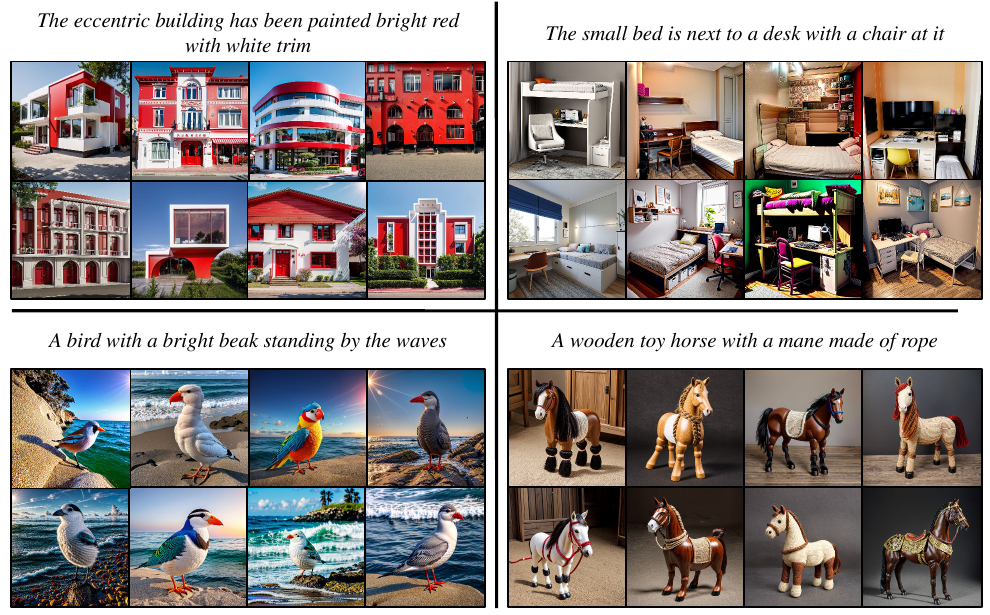}
    \caption[]{\textbf{Additional Qualitative Results.} Additional LoRAverse outputs showcasing diverse image sets generated by retrieving various LoRA models while maintaining relevance to the user prompts.
}\label{fig:additional-qualitative-results}
\end{figure*}

\begin{figure*}[]
    \centering
    \includegraphics[width=0.88\linewidth]{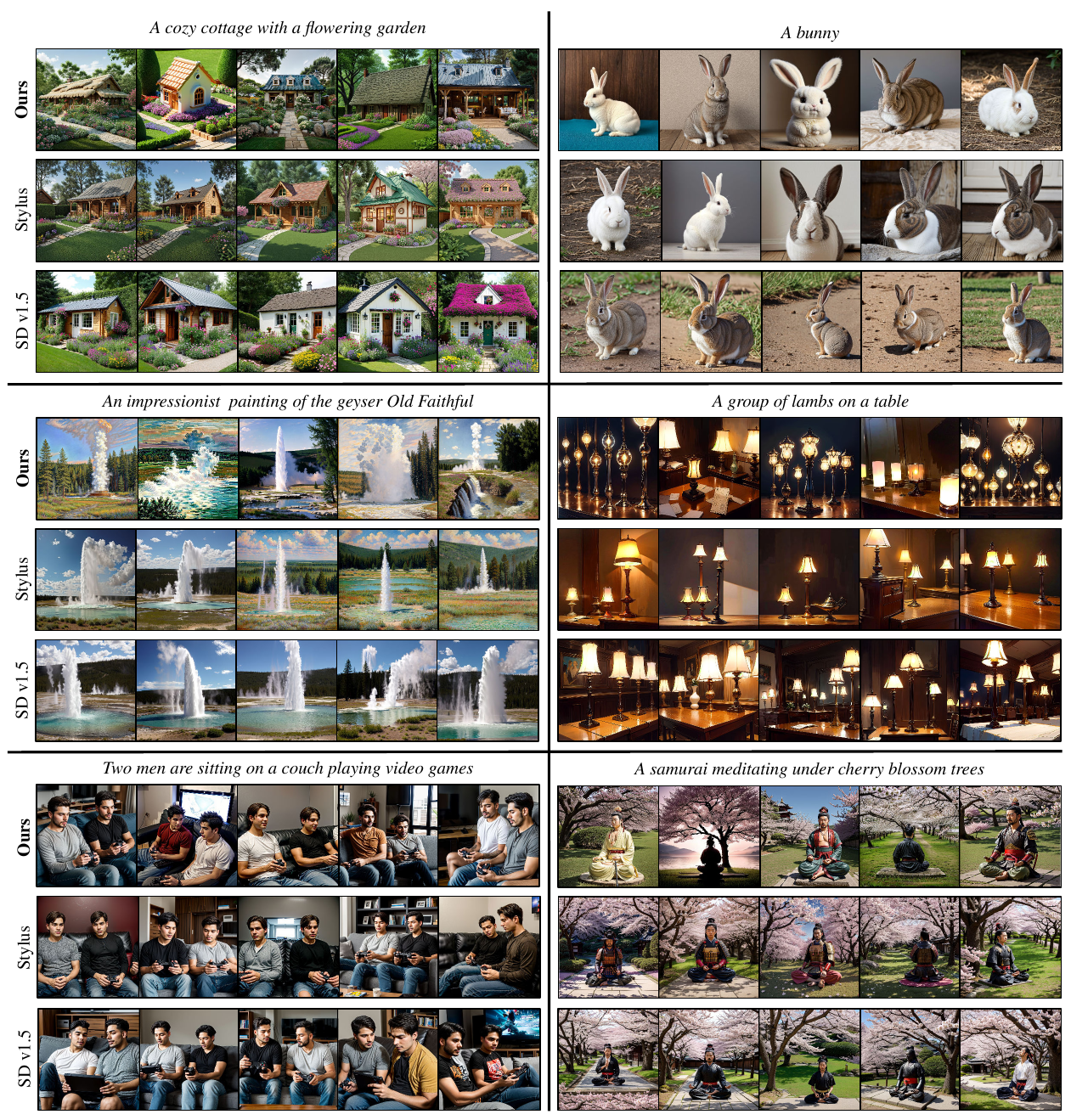}
    \caption[]{\textbf{Additional Qualitative Comparison.} Additional outputs generated by LoRAverse, Stylus, and SD v1.5 demonstrating diverse image sets.
}\label{fig:additional-qualitative-comparison}
\end{figure*}

Additional qualitative results are in Fig. \ref{fig:additional-qualitative-results} and Fig. \ref{fig:additional-qualitative-comparison}, demonstrating LoRAverse’s effectiveness in generating diverse and relevant image sets. Our approach balances diversity with image-text alignment, thanks to the clustering-based retrieval process. By selecting adapters from the clusters, our method ensures the generated images reflect both the prompt’s content and stylistic variety, offering a richer interpretation compared to other methods.

\begin{figure}[]
    \centering
    \includegraphics[width=\linewidth]{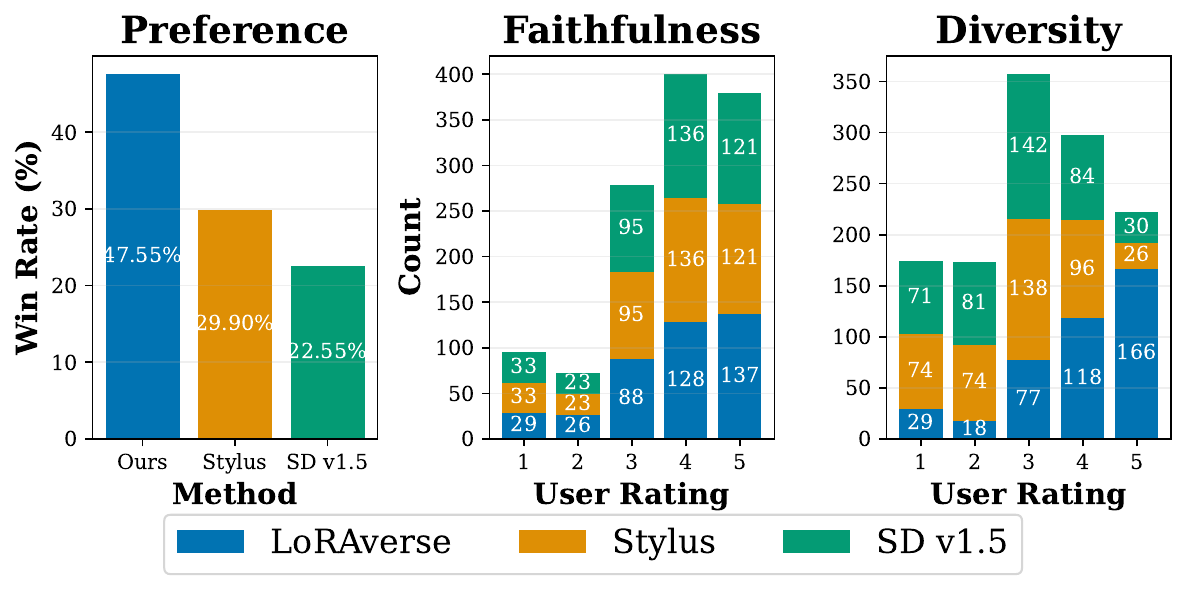}
    \caption[]{\textbf{User Study.} Users were asked between LoRAverse, Stylus, and SD v1.5 which produced preferred outputs, and also to rate the faithfulness and diversity of outputs on a scale of 1 to 5. LoRAverse exhibits both strong desirability and outperforms in diversity.
}\label{fig:user-study}
\end{figure}

\section{Additional Quantitative Results}
\label{sup:quantitative-results}

To contextualize LoRAverse against stronger yet straightforward alternatives, we add three baselines: RTop‑100 and RTop‑500, which rank all LoRA models by global prompt similarity and randomly sample $K=8$ from the top 100 or 500, and Random, which samples from the full pool. As summarized in Table \ref{tab:additional-quantitave-comparison}, LoRAverse matches RTop‑100 in CLIP alignment (24.94 vs.\ 24.99) while achieving higher diversity on all metrics and lower image‑to‑image similarity. The advantage widens for RTop‑500 and Random, whose looser sampling selects more off‑topic adapters, leading to lower CLIP scores. These results confirm that the proposed submodular retrieval offers a superior balance between diversity and text‑adherence.

\section{Ablation Study Over the Number of Clusters and Concepts}
\label{sup:cluster-concept-ablation}

We evaluated LoRAverse's sensitivity to the number of clusters formed per concept (5, 10, 25) and the number of concepts extracted from the prompt (1–3). Varying the size of the cluster effectively leaves all metrics stable, that is, it changes all metrics by less than 1.5 points and changes the pairwise similarity by less than 0.02, showing no consistent trend (see Table \ref{tab:additional-quantitave-comparison}). This indicates LoRAverse is robust to these settings.

\begin{table}[t]
\setlength{\tabcolsep}{2pt}
\footnotesize
\centering
\begin{tabular}{lcccc}
\hline
& CLIP ($\uparrow$) & TCE ($\uparrow$) & TIE ($\uparrow$) & I2I ($\downarrow$) \\ 
\hline
\textbf{LoRAverse} & 24.94 (\textcolor{green}{14.6\%}) & \textbf{25.72} (\textcolor{green}{2.3\%}) & \textbf{43.04} (\textcolor{green}{0.5\%}) & \textbf{0.719} (\textcolor{green}{-5.8\%}) \\
SD v1.5 & \textbf{25.29} (\textcolor{green}{16.2\%}) & 23.65 (\textcolor{red}{-6.0\%}) & 41.86 (\textcolor{red}{-2.3\%}) & 0.774 (\textcolor{red}{1.4\%}) \\
RTop-100 & 24.99 (\textcolor{green}{14.8\%}) & 25.16 (0.0\%) & 42.40 (\textcolor{red}{-1.3\%}) & 0.740 (\textcolor{green}{-3.0\%})  \\
RTop-500 & 24.48 (\textcolor{green}{12.5\%}) & 25.05 (\textcolor{red}{-0.4\%}) & 42.35 (\textcolor{red}{-1.1\%}) & 0.739 (\textcolor{green}{-3.2\%}) \\
Random & 21.76 & 25.15 & 42.84 & 0.763 \\
\hline
5-Cluster & 24.55 & 26.10 & 43.22 & \textbf{0.716} \\
10-Cluster & \textbf{24.94} & 25.72 & 43.04 & 0.719 \\
25-Cluster & 24.45 & \textbf{26.39} & \textbf{43.41} & 0.707 \\
\hline
1-Concept & 24.73 & 23.27 & 40.25 & 0.772 \\
2-Concept & 24.87 & \textbf{24.02} & 40.49 & 0.763 \\
3-Concept & \textbf{24.92} & 23.54 & \textbf{41.61} & \textbf{0.762} \\
\hline
\end{tabular}
\caption{\textbf{Additional Quantitative Comparison.} Results from the additional metrics and the ablation study on the number of clusters and concepts.}
\label{tab:additional-quantitave-comparison}
\end{table}

\begin{figure*}[]
    \centering
    \includegraphics[width=0.88\linewidth]{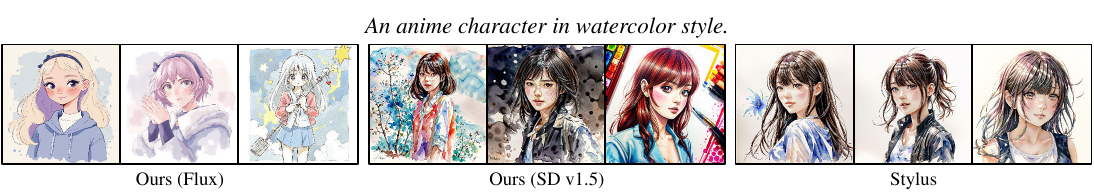}
    \caption[]{\textbf{Compatibility with Flux.} Additional qualitative outputs generated by Flux, LoRAverse, and Stylus.
}\label{fig:flux}
\end{figure*}

\begin{figure}[]
    \centering
    \includegraphics[width=\linewidth]{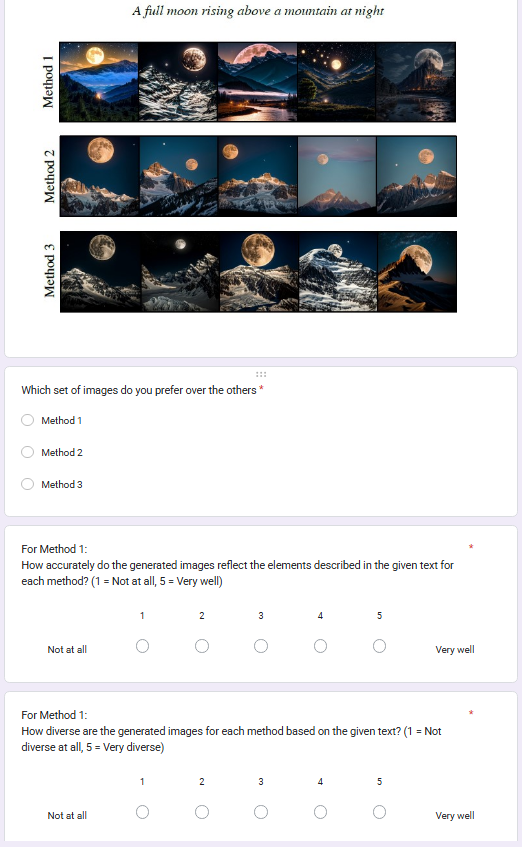}
    \caption[]{\textbf{Screenshot of the User Study.}
}\label{fig:user-study-ss}
\end{figure}

\section{Details of Concept Extractor}
\label{sup:concept-extractor-details}
A complete example prompt for the \textit{concept extractor} is provided in the first column of Table \ref{tab:component-prompts}, utilizing the \texttt{gpt-4o-mini} model. We design a structured prompt to help the LLM effectively identify distinct, orthogonal concepts while merging those that are semantically similar. This module is implemented using LangChain. \cite{Chase_LangChain_2022}.

\section{Details of Adapter Safety Checker}
\label{sup:safety-checker-details}

A complete input for the \textit{adapter safety checker} is provided in the second column of Table \ref{tab:component-prompts}, utilizing the \texttt{gpt-4o} model. This module ensures ethical integrity by filtering adapters associated with inappropriate, sexual, or anthropomorphic material. In large-scale text-to-image pipelines with diverse LoRA models, there exists a risk of retrieving adapters that could produce harmful outputs. The safety checker mitigates this by evaluating adapter descriptions and removing those violating content guidelines.

The checker is implemented through a structured prompt that helps the LLM identify inappropriate adapters unless explicitly requested by the user. It filters out two main types: (1) potentially sexual content, including nudity or explicit elements, and (2) anthropomorphic content, such as animal-inspired humanoids. This approach ensures only contextually appropriate adapters are selected.

The decision to filter anthropomorphic content stems from observations that these styles often blur fantasy and reality in problematic ways. These figures frequently appear in exaggerated forms that may contribute to objectification of human-like features. Excluding these adapters by default helps maintain more neutral outputs unless specifically requested.
Implemented using LangChain \cite{Chase_LangChain_2022}, the safety checker introduces computational overhead but remains a practical component. This approach balances creative diversity with safety guards, reducing the risk of generating unsafe outputs while preserving quality and diversity in the results.

\begin{table*}[h]
\centering
\scriptsize
\begin{tabular}{|p{0.45\textwidth}|p{0.45\textwidth}|}
\hline
\texttt{\textbf{System Prompt}} & \texttt{\textbf{System Prompt}} \\
You are a specialist to extract meaningful keywords and provide explanations for prompts designed for text-to-image models. Your task is to identify keywords or phrases that significantly impact the visual content generated by such models. 
& 
You are a specialized adapter content filter for text-to-image pipelines. Your primary task is to identify and filter out adapters that contain inappropriate content based on the following criteria:

1) Potentially Sexual Content: Descriptions referencing nudity, sexualized poses, or explicit elements.

2) Anthropomorphic Content: Descriptions featuring catgirls, animal-inspired humanoids, or similar anthropomorphic characters. \newline

Only exclude adapters containing these types of content unless the prompt explicitly mentions or requires them. You must carefully evaluate the descriptions of the adapters and flag those that violate the criteria. For each filtered adapter, provide its index and a concise explanation for why it was flagged. Your output should strictly follow the specified JSON format provided in the main prompt, listing the flagged adapters along with their explanations.
\\
\hline
\texttt{\textbf{User Prompt}} & \texttt{\textbf{User Prompt}} \\
You are a keyword extractor specialized in identifying concepts and their descriptive adjectives from prompts used in text-to-image models. Your task is to detect key concepts (keywords with their adjectives) that significantly influence the content, style, or composition of the generated image. Ensure that the extracted concepts are relevant to the visual elements of the scene and describe distinct entities or attributes. Do not split related concepts that describe a single entity. Avoid extracting verbs describing actions. \newline

Key Instructions:

1) Extract descriptive concepts: Always extract keywords that include any relevant descriptive adjectives or qualifiers that modify them in the prompt. If adjectives or modifiers are provided, ensure they are part of the concept, as they provide crucial detail to the entity. For example:

\qquad - For the prompt "a large, ancient stone castle," return the keyword "large, ancient stone castle" rather than just "castle." The adjectives "large," "ancient," and "stone" provide important context and description that shape the entity's appearance and character.
    
\qquad - For the prompt "a vibrant painting of a tropical sunset," return the keyword "vibrant painting of a tropical sunset," as "vibrant" enhances the concept of the painting and "tropical" adds a layer of specificity to the sunset.
    
2) Focus on impactful concepts: Keywords should describe specific entities, concepts, styles, or attributes central to the generated image. Avoid overly general terms (e.g., "thing" or "place"). Ignore verbs describing an action."

3) Combine related concepts: If multiple keywords collectively describe a single entity, treat them as one unified concept. Do not split descriptive phrases or components that together define the same object, scene, or idea. If extracted keywords simply describe or qualify a specific entity, they must be merged into a single concept. For example:

\qquad - In the prompt "a photo of San Francisco's Golden Gate Bridge," the entire phrase should be treated as one concept, "a photo of San Francisco's Golden Gate Bridge," because all parts describe a single entity—the photo.
    
\qquad - In the prompt "a picture of some food on the plate," the entire phrase should be treated as one concept, "a picture of some food on the plate," because all parts collectively describe the picture and its content.
    
4) Provide explanations: For each keyword, provide a concise explanation of why it was chosen, focusing on its role in shaping the image's appearance, theme, or composition.

5) Stick to the specified format: The output must follow the JSON format provided below within $<$output\_format$>$ and $<$/output\_format$>$ tags. \newline

$<$output\_format$>$

[

\qquad  \{
  
\qquad\qquad    "explanation": "[Reason for selecting this keyword]",
    
\qquad\qquad    "keyword": "[Keyword identified]"
    
\qquad  \},
  
\qquad  \{
  
\qquad\qquad    "explanation": "[Reason for selecting this keyword]",
    
\qquad\qquad    "keyword": "[Keyword identified]"
    
\qquad  \}, ...
  
]

$<$/output\_format$>$ \newline

Extract and explain keywords for the following prompt enclosed within $<$prompt$>$ and $<$/prompt$>$ tags:

$<$prompt$>$

\{\texttt{prompt}\}

$<$/prompt$>$ \newline

Provide your response in the JSON format specified above without the $<$output\_format$>$ and  ```json tags. 
&  
You are an adapter content filter for text-to-image models. Adapters are plug-in modules that enhance image quality in specific domains, as described in their descriptions. Your task is to identify adapters that are inappropriate for use in the context of the provided prompt and keyword. Specifically, filter out adapters whose descriptions contain the following types of content, unless the prompt explicitly mentions them:

1) Potentially Sexual Content: Descriptions referencing nudity, sexualized poses, or other explicit elements.

2) Anthropomorphic Content: Descriptions featuring catgirls, animal-inspired humanoids, or similar anthropomorphic characters. \newline

You are provided with adapter descriptions, each marked with an index value. Your response should list the indices of the inappropriate adapters, along with a brief explanation for why they were filtered. \newline

Key Instructions:

1) Filter Content Types: Remove adapters containing the specified types of content unless the prompt directly mentions or explicitly requires them.

2) Provide Explanations: For each filtered adapter, include a concise explanation of why it was flagged, referencing the content type(s) it violates.

3) Follow the output format: Use the JSON format provided below, enclosed within $<$output\_format$>$ and $<$/output\_format$>$ tags. \newline

$<$output\_format$>$

[

\qquad  \{

\qquad\qquad    "explanation": "[Reason for why this adapter is inappropriate]"
    
\qquad\qquad    "id": "[id of the inappropriate adapter]"
  
\qquad  \},
  
\qquad  \{
  
\qquad\qquad    "explanation": "[Reason for why this adapter is inappropriate]"
    
\qquad\qquad    "id": "[id of the inappropriate adapter]"
  
\qquad  \}, ...

]

$<$/output\_format$>$ \newline

Task Section:

Filter the inappropriate adapters for the provided keyword and prompt enclosed below:

$<$keyword\_and\_prompt$>$

Keyword: \{\texttt{keyword}\}

Prompt: \{\texttt{prompt}\}

$<$/keyword\_and\_prompt$>$ \newline

The adapter descriptions to be filtered are enclosed below:

$<$adapter\_descriptions$>$

\{\texttt{adapter\_descriptions\_str}\}

$<$/adapter\_descriptions$>$ \newline

Provide your response in the JSON format specified above without the  $<$output\_format$>$ and ```json tags.
\\
\hline
\end{tabular}
\caption{Complete prompts for the \textit{concept extractor} and \textit{adapter safety checker} respectively.}
\label{tab:component-prompts}
\end{table*}

\section{Details of VLM-as-a-Judge}
\label{sup:vlm-as-a-judge}

The complete prompt used for evaluating the diversity, quality, and textual alignment of the image sets with \texttt{gpt-4o} is outlined in Table \ref{tab:vlm-as-a-judge-prompts}. Our methodology involves presenting three distinct image sets and using multi-turn prompting techniques to differentiate between them. Each set is numbered sequentially, and the VLM is asked to evaluate their diversity, quality, and textual alignment. The prompt includes a structured rubric, clear instructions, reminders, and example model outputs. To quantify the metrics, the model uses Chain-of-Thought reasoning to assign scores ranging from 0 to 2, based on methodologies from Stylus \cite{dunlap2024describingdifferencesimagesets, wei2023chainofthoughtpromptingelicitsreasoning, yao2023treethoughtsdeliberateproblem}.

\section{Details of Clustering Implementation}
\label{sup:clustering}

To enhance reproducibility, we provide a detailed overview of our clustering methodology. We use the BERTopic \cite{grootendorst2022bertopic} framework with UMAP \cite{mcinnes2020umapuniformmanifoldapproximation} and HDBSCAN \cite{Malzer_2020} for adapter grouping. The UMAP configuration is set to 5 neighbors, 5 components, and cosine distance metric, prioritizing local structure preservation while reducing dimensionality efficiently. For clustering, HDBSCAN is configured with a minimum cluster size of 3, allowing for the formation of micro-clusters. We use euclidean distance in the low-dimensional space and enable prediction data to support unseen data inference. The number of clusters is not predetermined but emerges dynamically from the density-based clustering, ensuring adaptability across datasets. Additionally, we set the number of topics to 10. These design choices capture the underlying structure of LoRA models.

\section{Details of User Study}
\label{sup:user-study}

For each prompt, we presented a set of 5 images for LoRAverse, Stylus, and SD v1.5 and asked 3 questions:

\begin{itemize}
    \item \textbf{Question 1 - Preference}: "Which set of images do you prefer over the others"
    \item \textbf{Question 2 - Faithfulness}: "For [GIVEN METHOD]:
    How accurately do the generated images reflect the elements described in the given text for each method? (1 = Not at all, 5 = Very well)"
    \item \textbf{Question 3 - Diversity}: "For [GIVEN METHOD]:
How diverse are the generated images for each method based on the given text? (1 = Not diverse at all, 5 = Very diverse)"
\end{itemize}

Fig. \ref{fig:user-study} shows the win-rate (proportion of times a user chose a method as their preferred), along with the distribution of user ratings for each method for faithfulness and diversity. We observe that while users gave relatively similar ratings for the image faithfulness, there are significantly more users rating LoRAverse with higher diversity. A screenshot of the survey for a sample prompt and question can be seen in Fig. \ref{fig:user-study-ss}.

\section{Compatibility with Other Text-to-Image Models}
\label{sup:compatibility}

LoRAverse is inherently backbone‑agnostic because it represents each adapter solely by its CLIP embedding, the submodular retrieval objective--and thus the entire pipeline--remains unchanged when the underlying text‑to‑image model is swapped. To validate this claim, we indexed 300 publicly released Flux-compatible LoRA adapters from Hugging Face \cite{hf_website}, created their embeddings, and ran our selection algorithm using Flux.1 backbone (see Fig. \ref{fig:flux}).

\begin{table*}[h]
\centering
\tiny
\begin{tabular}{|p{0.3\textwidth}|p{0.3\textwidth}|p{0.3\textwidth}|}
\hline
\texttt{\textbf{System Prompt}} & \texttt{\textbf{System Prompt}} & \texttt{\textbf{System Prompt}} \\
You are a precise and objective Photoshop expert tasked with evaluating the diversity of three given image sets. Your role is to analyze and score the diversity of these sets based on predefined criteria. You must provide a clear decision on which image set is more diverse, along with detailed explanations for your reasoning. Your assessment should be factual, concise, and unbiased, following the specified JSON format. \newline

Scoring Criteria:

Diversity scores are assigned as follows:

- 2 (Very diverse): The image set displays significant variation across themes and main subjects.

- 1 (Somewhat diverse): The set shows some diversity but lacks variation in either theme interpretation or main subjects.

- 0 (Not diverse): The set contains minimal or no variation in both theme interpretation and main subjects. \newline

Diversity Evaluation Breakdown:

You will assess diversity based on two key aspects:

1) Theme Interpretation: The theme should exhibit multiple interpretations. For example, if the theme is "It's raining cats and dogs", a diverse set should include both literal (cats and dogs falling from the sky) and figurative (heavy rain) representations. If the set only includes images of heavy rain or only of animals, it should receive a score of 1 instead of 2.

2) Main Subject: The diversity score should reflect changes in the primary subject of the images. For example, a set containing images of apples and children dressed as apples is more diverse than a set with only children dressed as apples. A set with varying focal points across different images should receive a higher diversity score.

&

You are a precise and objective Photoshop expert tasked with evaluating the composition quality of three given image sets. Your role is to analyze and score the quality of these sets based on predefined criteria. You must provide a clear decision on which image set has higher quality, along with detailed explanations for your reasoning. Your assessment should be factual, concise, and unbiased, following the specified JSON format. \newline

Scoring Criteria:

Compositional quality scores are assigned as follows:

- 2 (Very quality): The image set displays high quality.

- 1 (Somewhat quality): The image set is visually aesthetic but has elements with distortion, missing, or extra features.

- 0 (Not quality): The set contains minimal or no quality. \newline

Quality Evaluation Breakdown:

You will assess quality based on three key aspects:

1) Clarity:  Score 0 if the image is blurry, poorly lit, or has poor composition with objects obstructing each other.

2) Disfigured Parts: This applies to both body parts of humans/animals and objects like motorcycles. Score 0 if parts are severely disfigured such as fingers showing lips and teeth warped in. Score 1 for minor anatomical errors like a hand with 6 fingers.

3) Detail:  Score 0 for the appearance inconsistent with the environment. Score 1 for acceptable but basic detail such as monochrome and flat surfaces. Score 2 for rich, realistic detail like a sailboat showing dynamic ripples and ornate patterns.

&

You are a precise and objective Photoshop expert tasked with evaluating the textual alignment of three given image sets based on the provided prompt. Your role is to analyze and score the textual alignment of these sets according to the following criteria. You must provide a clear \
decision on which image set aligns best with the prompt, along with detailed explanations for your reasoning. Your assessment should be factual, concise, and unbiased, following the specified JSON format. \newline

Scoring Criteria:

Textual alignment scores are assigned as follows:

- 2 (Fully aligned): The image set displays high textual alignment with the prompt.

- 1 (Somewhat aligned): The set incorporates part of the theme but not all elements are correctly represented.

- 0 (Not aligned): The set contains minimal or no textual alignment with the prompt. \newline

Here are some examples:

- If the prompt is "shoes" and an image depicts a sock, the score would be 0 (not aligned).

- If the prompt is "shoes without laces" but the image shows shoes with laces, the score would be 1 (somewhat aligned).

- If the prompt is "a concert without fans," but an image includes fans, select the set with fewer fans. This would be scored based on the image set that most closely matches the prompt.
\\
\hline
\texttt{\textbf{User Prompt}} & \texttt{\textbf{User Prompt}} & \texttt{\textbf{User Prompt}} \\
This is one of the image sets. Please reply 'ACK'.

$<$image\_set\_1$>$ & This is one of the image sets. Please reply 'ACK'.

$<$image\_set\_1$>$ & This is one of the image sets. Please reply 'ACK'.

$<$image\_set\_1$>$ \\
\hline
\texttt{\textbf{Assistant}} & \texttt{\textbf{Assistant}} & \texttt{\textbf{Assistant}} \\
ACK & ACK & ACK \\
\hline
\texttt{\textbf{User Prompt}} & \texttt{\textbf{User Prompt}} & \texttt{\textbf{User Prompt}} \\
This is one of the image sets. Please reply 'ACK'.

$<$image\_set\_2$>$ & This is one of the image sets. Please reply 'ACK'.

$<$image\_set\_2$>$ & This is one of the image sets. Please reply 'ACK'.

$<$image\_set\_2$>$ \\
\hline
\texttt{\textbf{Assistant}} & \texttt{\textbf{Assistant}} & \texttt{\textbf{Assistant}} \\
ACK & ACK & ACK \\
\hline
\texttt{\textbf{User Prompt}} & \texttt{\textbf{User Prompt}} & \texttt{\textbf{User Prompt}} \\
This is one of the image sets. Please reply 'ACK'.

$<$image\_set\_3$>$ & This is one of the image sets. Please reply 'ACK'.

$<$image\_set\_3$>$ & This is one of the image sets. Please reply 'ACK'.

$<$image\_set\_3$>$ \\
\hline
\texttt{\textbf{Assistant}} & \texttt{\textbf{Assistant}} & \texttt{\textbf{Assistant}} \\
ACK & ACK & ACK \\
\hline
\texttt{\textbf{User Prompt}} & \texttt{\textbf{User Prompt}} & \texttt{\textbf{User Prompt}} \\
Rate the diversity of the three provided image sets using the scoring criteria above. For each group, assign each set a diversity score along with a detailed explanation in the following JSON output format: \newline

JSON Output Format:
[
    
\qquad    \{
        
\qquad\qquad        "image\_set\_1\_explanation": \#Your detailed evaluation of the diversity in Image Set 1\#,
        
\qquad\qquad        "image\_set\_1\_score": \#2, 1, or 0\#
    
\qquad    \},
    
\qquad    \{
        
\qquad\qquad        "image\_set\_2\_explanation": \#Your detailed evaluation of the diversity in Image Set 2.\#,
        
\qquad\qquad        "image\_set\_2\_score": \#2, 1, or 0\#
    
\qquad    \},
    
\qquad    \{
     
\qquad\qquad        "image\_set\_3\_explanation": \#Your detailed evaluation of the diversity in Image Set 3.\#,
        
\qquad\qquad        "image\_set\_3\_score": \#2, 1, or 0\#
    
\qquad    \},
    
\qquad    \{
    
\qquad\qquad        "preference\_explanation": \#Your reasoning for choosing the more diverse set.\#,
        
\qquad\qquad        "choice": \#IMAGE\_SET\_1, IMAGE\_SET\_2, or IMAGE\_SET\_3\#
    
\qquad    \}

] \newline 

I will make my own judgement using your results, your response is just an opinion as part of a rigorous process. I provide additional requirements below:

- Do not forget to reward different main subjects in the diversity score.

- You must pick a group for "More Diversity," neither is not an option.

- If the decision is close, make a choice and clarify your reasoning. \newline

Provide your response directly in the specified JSON format without ```json tags.

& 

Rate the quality of the three provided image sets using the scoring criteria above. For each group, assign each set a quality score along with a detailed explanation in the following JSON output format: \newline

JSON Output Format:
[
    
\qquad    \{
        
\qquad\qquad        "image\_set\_1\_explanation": \#Your detailed evaluation of the quality in Image Set 1\#,
        
\qquad\qquad        "image\_set\_1\_score": \#2, 1, or 0\#
    
\qquad    \},
    
\qquad    \{
        
\qquad\qquad        "image\_set\_2\_explanation": \#Your detailed evaluation of the quality in Image Set 2.\#,
        
\qquad\qquad        "image\_set\_2\_score": \#2, 1, or 0\#
    
\qquad    \},
    
\qquad    \{
     
\qquad\qquad        "image\_set\_3\_explanation": \#Your detailed evaluation of the quality in Image Set 3.\#,
        
\qquad\qquad        "image\_set\_3\_score": \#2, 1, or 0\#
    
\qquad    \},
    
\qquad    \{
    
\qquad\qquad        "preference\_explanation": \#Your reasoning for choosing the higher quality set.\#,
        
\qquad\qquad        "choice": \#IMAGE\_SET\_1, IMAGE\_SET\_2, or IMAGE\_SET\_3\#
    
\qquad    \}

] \newline

I will make my own judgement using your results, your response is just an opinion as part of a rigorous process. I provide additional requirements below:

- You must pick a group for "Better Quality," neither is not an option.

- If the decision is close, make a choice and clarify your reasoning. \newline

Provide your response directly in the specified JSON format without ```json tags.

&

Rate the textual alignment of the three provided image sets using the scoring criteria above. For each group, assign each set a textual alignment score along with a detailed explanation in the following JSON output format:

JSON Output Format:
[
    
\qquad    \{
        
\qquad\qquad        "image\_set\_1\_explanation": \#Your detailed evaluation of the textual alignment in Image Set 1\#,
        
\qquad\qquad        "image\_set\_1\_score": \#2, 1, or 0\#
    
\qquad    \},
    
\qquad    \{
        
\qquad\qquad        "image\_set\_2\_explanation": \#Your detailed evaluation of the textual alignment in Image Set 2.\#,
        
\qquad\qquad        "image\_set\_2\_score": \#2, 1, or 0\#
    
\qquad    \},
    
\qquad    \{
     
\qquad\qquad        "image\_set\_3\_explanation": \#Your detailed evaluation of the textual alignment in Image Set 3.\#,
        
\qquad\qquad        "image\_set\_3\_score": \#2, 1, or 0\#
    
\qquad    \},
    
\qquad    \{
    
\qquad\qquad        "preference\_explanation": \#Your reasoning for choosing the better textual alignment set.\#,
        
\qquad\qquad        "choice": \#IMAGE\_SET\_1, IMAGE\_SET\_2, or IMAGE\_SET\_3\#
    
\qquad    \}

] \newline

Prompt:

{prompt} \newline

I will make my own judgement using your results, your response is just an opinion as part of a rigorous process. I provide additional requirements below:

- You must pick a group for "Better Textual Alignment," neither is not an option.

- If the decision is close, make a choice and clarify your reasoning. \newline

Provide your response directly in the specified JSON format without ```json tags. \\
\hline
\end{tabular}
\caption{Complete prompts for the VLM-as-a-Judge to evaluate the image diversity, quality, and textual alignment respectively.}
\label{tab:vlm-as-a-judge-prompts}
\end{table*}

\end{document}